\documentclass[10pt,twocolumn,letterpaper]{article}

\usepackage{cvpr}              

\usepackage{graphicx}
\usepackage{amsmath}
\usepackage{amssymb}
\usepackage{booktabs}
\usepackage{amsmath}
\usepackage{algpseudocode}
\usepackage{algorithm}
\usepackage{tabularx}

\usepackage{color,soul}
\usepackage{bbding}
\soulregister\cite7
\soulregister\ref7
\soulregister\pageref7

\newcommand\mdoubleplus{\mathbin{+\mkern-10mu+}}

\makeatletter
\robustify\@latex@warning@no@line
\makeatother
\usepackage{authblk}

\usepackage[pagebackref,breaklinks,colorlinks]{hyperref}

\usepackage[capitalize]{cleveref}
\crefname{section}{Sec.}{Secs.}
\Crefname{section}{Section}{Sections}
\Crefname{table}{Table}{Tables}
\crefname{table}{Tab.}{Tabs.}

\def\cvprPaperID{5609} 
\def\confName{CVPR}
\def\confYear{2023}

\begin{document}

\title{2PCNet: Two-Phase Consistency Training for Day-to-Night \\ Unsupervised Domain Adaptive Object Detection}

\setlength{\affilsep}{0em}
\author[1,2]{Mikhail Kennerley}
\author[2]{Jian-Gang Wang}
\author[1]{Bharadwaj Veeravalli}
\author[1]{Robby T. Tan}
\affil[1]{National University of Singapore, Department of Electrical and Computer Engineering} 
\affil[2]{Institute for Infocomm Research, A*STAR} 
\affil[ ]{\tt\small mikhailk@u.nus.edu, jgwang@i2r.a-star.edu.sg, elebv@nus.edu.sg, robby.tan@nus.edu.sg}

\maketitle


\begin{abstract}

Object detection at night is a challenging problem due to the absence of night image annotations. 
Despite several domain adaptation methods, achieving high-precision results remains an issue. 
False-positive error propagation is still observed in methods using the well-established student-teacher framework, particularly for small-scale and low-light objects. 
This paper proposes a two-phase consistency unsupervised domain adaptation network, 2PCNet, to address these issues. 
The network employs high-confidence bounding-box predictions from the teacher in the first phase and appends them to the student's region proposals for the teacher to re-evaluate in the second phase, resulting in a combination of high and low confidence pseudo-labels. 
The night images and pseudo-labels are scaled-down before being used as input to the student, providing stronger small-scale pseudo-labels. 
To address errors that arise from low-light regions and other night-related attributes in images, we propose a night-specific augmentation pipeline called NightAug. 
This pipeline involves applying random augmentations, such as glare, blur, and noise, to daytime images.
Experiments on publicly available datasets demonstrate that our method achieves superior results to state-of-the-art methods by 20\%, and to supervised models trained directly on the target data. \footnote{\url{www.github.com/mecarill/2pcnet}}

\end{abstract}


\section{Introduction}
\label{sec:intro}

\begin{figure}[t]
\centering

\includegraphics[width=0.47\textwidth]{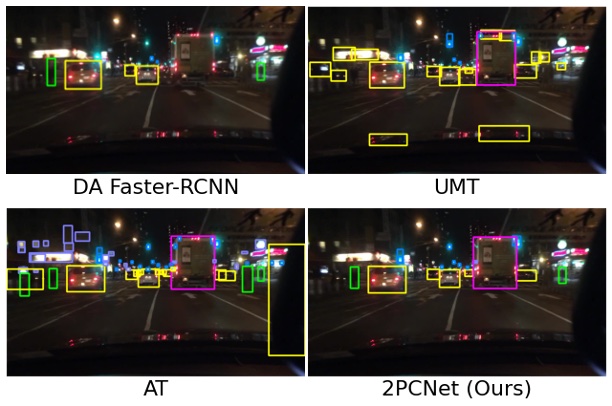}
\caption{
	Qualitative results of state-of-the-art DA methods, DA Faster-RCNN \cite{chen2018domain}, UMT \cite{deng2021unbiased}, Adaptive Teacher (AT) \cite{li2022cross} and our method 2PCNet on the BDD100K \cite{bdd100k} dataset. 
	Unlike the SOTA methods, our method is able to detect dark and small scale objects with minimal additional false positive predictions.
}
\label{fig:bdd100kqual}
\end{figure}

Nighttime object detection is critical in many applications. 
However, the requirement of annotated data by supervised methods is impractical, since night data with annotations is few, and supervised methods are generally prone to overfitting to the training data. 
Among other reasons, this scarcity is due to poor lighting conditions which makes nighttime images hard to annotate.
Hence,  methods that do not assume the availability of the annotations are more advantageous.
Domain adaptation (DA) is an efficient solution to this problem by allowing the use of readily available annotated source daytime datasets.

A few domain adaptation methods have been proposed, e.g.,  adversarial learning which uses image and instance level classifiers \cite{chen2018domain} and similar concepts \cite{saito2018strong, Xu_2020_CVPR}. 
However, these methods isolate the domain adaptation task purely towards the feature extractor, and suppress features of the target data for the sake of domain invariance.
Recent unsupervised domain adaptation methods exploit the student-teacher framework (e.g.~\cite{deng2021unbiased,li2022cross,He_2022_CVPR,8953637}).
Since the student initially learns from the supervised loss, there is a bias towards the source data.
Augmentation \cite{deng2021unbiased, He_2022_CVPR} and adversarial learning \cite{li2022cross} have been proposed  to address this problem. 
Unfortunately, particularly for day-to-night unsupervised domain adaptation,  these methods suffer from a large number of inaccurate pseudo-labels produced by the teacher. 
In our investigation, the problem is notably due to insufficient knowledge of small scale features in the nighttime domain, which are then propagated through the learning process between the teacher and student, resulting in poor object detection performance.
%

%

To address the problem, in this paper, we present 2PCNet, a two-phase consistency unsupervised domain adaptation network for nighttime object detection.
Our 2PCNet merges the bounding-boxes of highly-confident pseudo-labels, which are predicted in phase one, together with regions proposed by the student's region proposal network (RPN).  The merged proposals are then used by the teacher to generate a new set of pseudo-labels in phase two. 
This provides a combination of high and low confidence pseudo-labels.
These pseudo-labels are then matched with predictions generated by the student. 
We can then utilise a weighted consistency loss to ensure that a higher weightage of our unsupervised loss is based on stronger pseudo-labels, yet allow for weaker pseudo-labels to influence the training.

Equipped with this two-phase strategy, we address the problem of errors from small-scale objects.
We devise a student-scaling technique, where night images and their pseudo-labels for the student are deliberately scaled down. 
In order to generate accurate pseudo-labels, images to the teacher remain at their full scale.
This results in the pseudo-labels of larger objects, which are easier to predict, to be scaled down to smaller objects, allowing for an increase in small scale performance of the student.

Nighttime images suffer from multiple complications not found in daytime scenes such as dark regions, glare, prominent noise, prominent blur, imbalanced lighting, etc. 
All these cause a problem, since the student, which was trained on daytime images, is much more biased towards the daytime domain's characteristics. 
To mitigate this problem, we propose NightAug, a set of random nighttime specific augmentations. 
NightAug includes adding artificial glare, noise, blur, etc. that mimic the night conditions to daytime images.
With NightAug we are able to reduce the bias of the student network towards the source data without resulting to adversarial learning or compute-intensive translations.
Overall, using 2PCNet, we can see the qualitative improvements of our result in Figure \ref{fig:bdd100kqual}.
In summary, the contributions of this paper are as follows:
\begin{itemize}
	\setlength{\itemsep}{1pt}
	\item 
	We present 2PCNet, a two-phase consistency approach for student-teacher learning. 2PCNet takes advantage of highly confident teacher labels augmented with less confident regions, which are proposed by the scaled student.
	This strategy produces a sharp reduction of the error propagation in the learning process.

	\item To address the bias of the student towards the source domain, we propose NightAug, a random night specific augmentation pipeline to shift the characteristics of daytime images toward nighttime.

	\item The effectiveness of our approach has been verified by comparing it with the state-of-the-art domain adaptation approaches. An improvement of +7.9AP(+20\%) and +10.2AP(26\%) over the SOTA on BDD100K and SHIFT has been achieved, respectively. 
	
\end{itemize}


\begin{figure*}[t]
	\centering
	\includegraphics[width=\textwidth, height=9cm]{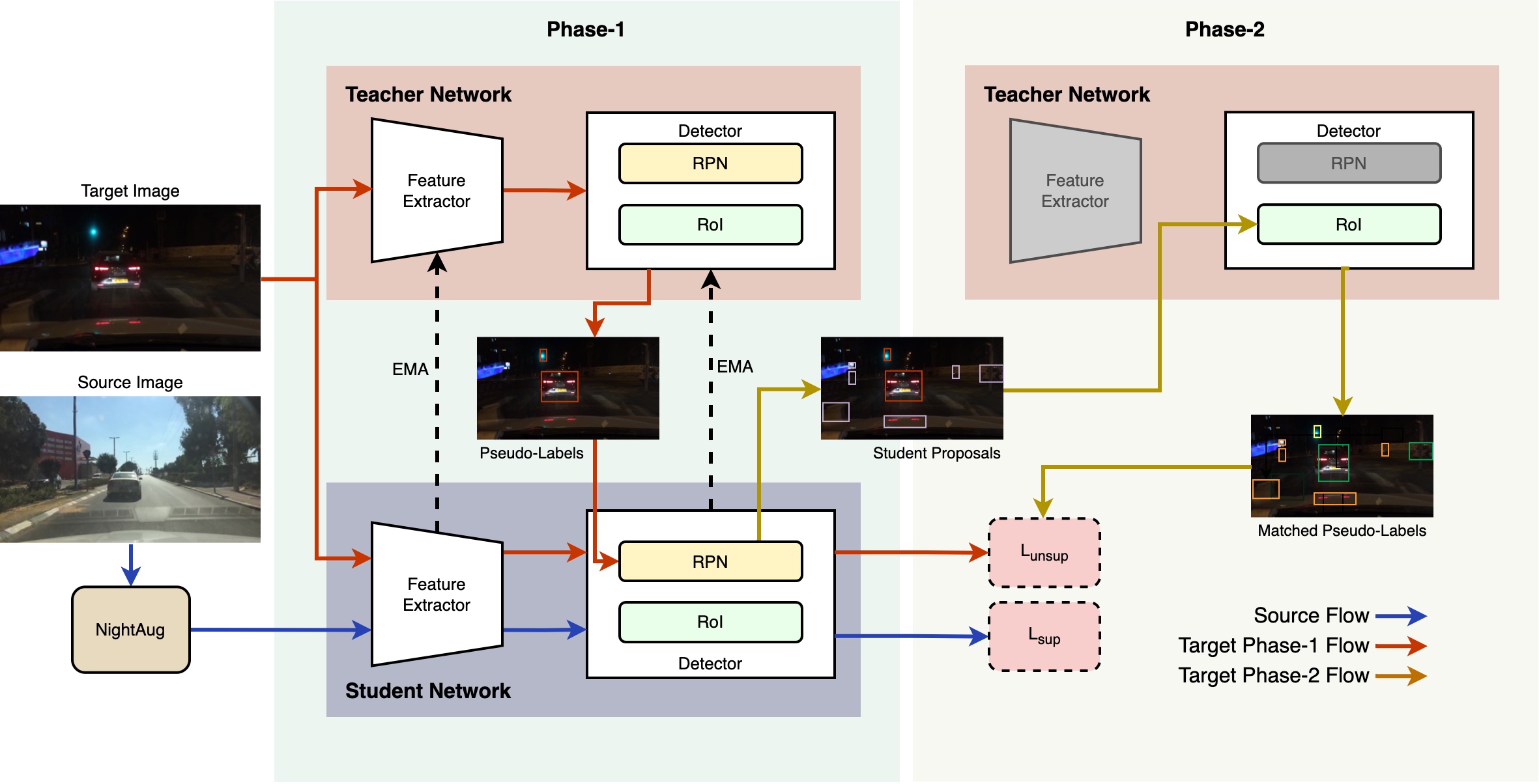}
	\caption{Overview of our proposed framework, 2PCNet. 2PCNet consists of: A student network is trained on both the labelled daytime image, which has been augmented with NightAug, and unlabelled nighttime images. A teacher network which is the exponential moving average (EMA) of the student and provides matched pseudo-labels for unsupervised loss. The match pseudo-labels are the predictions of the teacher (phase two) using the RPN proposals of the student, which in turn was guided by the high confidence pseudo-labels of the teacher (phase one).}
	\label{fig:arch}
\end{figure*}

\section{Related Work}

\vspace{-0.01cm}
\paragraph{Unsupervised Domain Adaptation (UDA)} Unsupervised domain adaptation aims to learn transferable features to reduce the discrepancy between a labelled source and unlabelled target domain. Previous works minimised the distance metric (MMD) \cite{long15,long16,long17} and considered intra-class and inter-class discrepancy \cite{kang2019,kang2020}. Adversarial feature learning involved adding an adversarial classifier to play the min-max game between the domain discriminator and feature extractors to generate a domain invariant feature map \cite{Tzeng_2017_CVPR,zhang2018coll,Wang_2020_CVPR}. These methods have been applied to image classification. Our work focuses on object detection, which is more complex as it involves identifying multiple bounding boxes and associated classes in each image.
\vspace{-0.3cm}
\paragraph{UDA for Object Detection}
Object detection with UDA is a recent challenge due to the complexities of identifying multiple objects in an image.
DA-Faster RCNN \cite{chen2018domain} integrated adversarial learning with image and instance level classifiers, and several approaches have been proposed to improve on this method by introducing scale-awareness \cite{chen2021sada}, class specific discriminators \cite{Xu2020ExploringCR}, and re-purposing the task-specific classifier as a discriminator \cite{daln2022}. 
The Mean Teacher (MT) framework \cite{antti2017MT} has been adopted in semi-supervised methods, such as UMT \cite{deng2021unbiased}, which incorporates CycleGAN \cite{CycleGAN2017} augmented images; AT \cite{li2022cross}, which combines the student-teacher framework with adversarial learning; and TDD \cite{He_2022_CVPR}, which uses dual student-teacher networks with style transfer.

\paragraph{Nighttime UDA} 
The majority of research on unsupervised domain adaptation (UDA) in nighttime scenarios has focused on semantic segmentation \cite{xuCDAdaCurriculumDomain2021, wuDANNetOneStageDomain2021,
SDV19,
lengyelZeroShotDayNightDomain2021,
gaoCrossDomainCorrelationDistillation2022,
dengNightLabDualLevelArchitecture2022,
d2ndata}.
Translation and style transformation techniques are commonly used to reduce the domain gap between the source and target domains in these methods \cite{dengNightLabDualLevelArchitecture2022,xuCDAdaCurriculumDomain2021,
wuDANNetOneStageDomain2021}. 
Some UDA-based techniques for nighttime also utilise paired-images to generate a shared feature space \cite{SDV19}, while others use an intermediate domain such as twilight to reduce the domain gap during unsupervised learning \cite{d2ndata}.

Nighttime tracking has also been investigated where adversarial transformers are used to close the domain gap \cite{yeUnsupervisedDomainAdaptation2022}. 
However, there is a gap in research when it comes to applying UDA techniques in the object detection task for nighttime scenarios.
Therefore, we explore the application of UDA techniques in object detection under low-light and nighttime conditions.
%

%

\section{Proposed Method}

 Let $\mathbf{D}_s$ be the daytime source data.
 $\mathbf{D}_s=\{I_s,C_s,B_s\}$, where the variables refer to the image, class label and bounding-box label, respectively. 
 Index $s$ indicates the daytime source. 
 The night target data is represented by $\mathbf{D}_t$, where $\mathbf{D}_t=\{I_t\}$ as we do not have the target labels available to us. 
 Index $t$ indicates the nighttime target.

The architecture of our 2PCNet is shown in Figure \ref{fig:arch}. 
Our 2PCNet consists of a student and a teacher network. 
The student is a multi-domain network trained on both labelled daytime images, augmented with NightAug, and unlabelled nighttime images. 
The teacher focuses on night images to produce pseudo-labels for the student and is the exponential moving average (EMA) of the student.
After an initial pretraining phase, the teacher begins producing pseudo-labels, which allows the student to initialise the feature extractor and detector. 

During each iteration, in phase one of 2PCNet, the teacher produces pseudo-labels from the night images. These pseudo-labels are filtered through a confidence threshold. This is to ensure only high-confidence pseudo-labels are given to the student. 
The bounding-boxes from the pseudo-labels are then combined with the region proposals generated by the student's RPN. 
The merged region proposals are then used to generate predictions from the student's RoI network.
In phase two, the teacher utilises the same merged region proposals to generate a matched set of pseudo-labels, where each pseudo-label has its corresponding prediction obtained from the student.  

As mentioned earlier, our student network is initialised by pretraining for a set number of iterations. This is done with supervised loss on the augmented daytime images:
\begin{equation}
\begin{split}
L_{\rm sup} = L_{\rm rpn}(B_s,I_s) + L_{\rm roi}(B_s,C_s,I_s),
\end{split}
\label{eq:sup}
\end{equation}
where $L_{\rm rpn}$ represents the loss from the RPN, which consists of an objectness and bounding-box regression loss.
$L_{\rm roi}$ represents the loss from the detector network, consisting of a classification and bounding-box regression loss.

Once the pretraining is completed, the student's weights are then transferred over to the teacher.
In the succeeding iterations, the teacher's weights are the exponential moving average (EMA) of the student's.  
The matched pseudo-labels generated by the teacher, $\{C_p^*,B_p^*\}$, are then used to guide the unsupervised loss, defined as:
\begin{equation}
\begin{split}
L_{\rm unsup} = L_{\rm rpn}^{\rm obj}(C_p^*;I_t) + L_{\rm cons}(C_p^*;I_t),
\end{split}
\label{eq:unsup}
\end{equation}
where $L_{\rm rpn}^{\rm obj}$ is the objectness loss of the RPN and $L_{\rm cons}$ is the weighted KL-Divergence loss from the predicted outputs which we will further explain in the next section.

\begin{figure}
	\centering
	\includegraphics[width=0.5\textwidth]{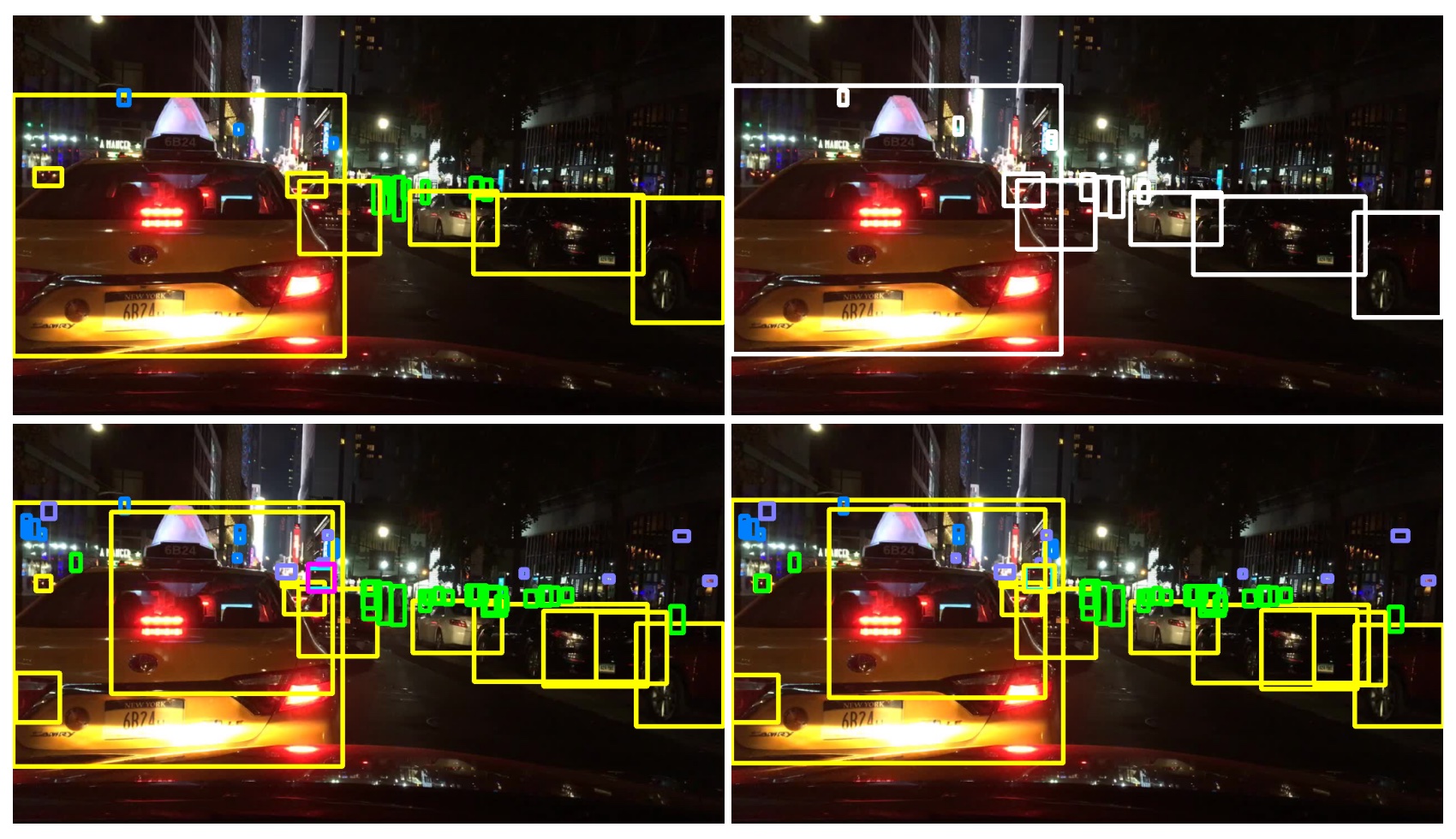}
	\caption{(Left to Right, Top to Bottom) Ground truth bounding boxes, bounding boxes predicted by the teacher with non-maximal suppression (NMS) and thresholding ($B_p$), bounding boxes predicted by the student ($B_{\rm student}$) which is guided by $B_p$, and the bounding boxes predicted by the teacher ($B_p^*$) for the consistency loss.}
	\label{fig:2pc}
\end{figure}

\subsection{Two-Phase Consistency}
Due to the large domain gap between daytime source images  and nighttime target images, the teacher is unable to produce high quality pseudo-labels.
This generally occurs in the whole scene, but particularly for regions with strong night characteristics, e.g., low-light, glare, uneven lighting, etc. 
The teacher produces confident pseudo-labels only for regions that share more similarities to the daytime, since it is biased towards the daytime domain.
This bias poses a problem for methods that employ a hard-threshold to filter pseudo-labels for categorical cross-entropy loss \cite{antti2017MT, deng2021unbiased,li2022cross}. 
The remaining pseudo-labels contain only easy samples with daytime attributes. 
Consequently, the student does not learn from harder (e.g. darker) areas.

%
%

As a result of minimal knowledge of the hard samples (i.e., areas with a high level of nighttime attributes), the teacher begins to predict highly confident yet incorrect pseudo-labels.
As the teacher provides these incorrect pseudo-labels to the student, a viscous cycle starts where the teacher in turn is updated with incorrect knowledge. Consequently, the error continues to propagate through training. 
In our case, these errors notably occur in dark/glare regions and as small scale objects.

To address the problem of error propagation, we design a two-phase approach that combines high confidence pseudo-labels together with their less confident counterparts. This combination allows for the high accuracy of confident-labels with the additional knowledge of less confident labels to be distilled onto the student.
In phase one, the unlabelled nighttime image, $I_t$, is used as an input for the teacher to generate pseudo-labels. These pseudo-labels are filtered with a threshold to retain only high-confidence pseudo-labels, $(C_p,B_p)$. 
The bounding-box of the pseudo-labels, $B_p$, is then used as an input to the student. $B_p$ is concatenated to the region proposals generated by the student RPN module:
\begin{equation}
\begin{split}
P^* = \text{RPN}_\text{student}(I_t) \mdoubleplus B_p,
\end{split}
\end{equation}
where $P^*$ is the combined region proposals, which are then used as an input to the student's RoI module to predict the classes, $C_{\rm student}$, and bounding-box, $B_{\rm student}$, of each region proposal.

Phase two begins by using the same combined region proposals, $P^*$, generated in phase one as an input to the teachers RoI module to generate a matched set of pseudo-labels:
\begin{equation}
\begin{split}
\{C_p^*, B_p^*\} = {\rm RoI}_{\rm teacher}(P^*).
\end{split}
\end{equation}

The difference between $C_p$ and $C_p^*$ is that $C_p^*$ is derived from the same region proposals as that of the student predictions $C_{\rm student}$. 
%
This allows us to compare $C_{\rm student}$ and $C_p^*$ directly:
\begin{equation}
\begin{split}
\{C_{\rm student}(n), B_{\rm student}(n)\}={\rm RoI}_{\rm student}(P^*(n)), \\ 
\{C_p^*(n), B_p^*(n)\} = {\rm RoI}_{\rm teacher}(P^*(n)),
\end{split}
\end{equation}
where $n=\{1,2,..,N\}$ and $N$ is the number of region proposals in $P^*$. 
This operation ensures that the knowledge of highly confident predictions generated by the teacher is distilled through to the student. In  addition, information from less confident predictions can also be learnt.
However, we are still required to penalise less confident samples and thus employ weighed KL-Divergence to be used as our consistency loss:
\begin{equation}
\begin{split}
L_{\rm cons} = \alpha \hspace{0.1cm} {\rm KL}(C_{\rm student},C_p^*),
\end{split}
\end{equation}
where $\alpha$ is the highest confidence of $C_p^*$ expressed as $\alpha = \max(C_p^*)$; ${\rm KL()}$ is the KL-divergence function.
Note that, pseudo-bounding boxes are not used to generate unsupervised loss, 
 as the confidence score of each pseudo-label represents the class information rather than the bounding box. 
The outputs of each segment of our two-phase approach are shown in Figure \ref{fig:2pc}.

\begin{algorithm}[t]
\caption{Single Augmentation - NightAug}\label{alg:cap}
\begin{algorithmic}
\State $\rm imgClean \gets img$
\If{$\rm randFloat\geq 0.5$} 
	\State $\rm randFloat \gets 0.8*randFloat + 0.2$
    \State $\rm img \gets augmentation(img,randval)$
    \State $\rm prob \gets 0.4$
    \While{$\rm randFloat\geq prob$}
    	\State $x \rm \gets randInt(img.shape[1],2)$
    	\State $y \rm \gets randInt(img.shape[2],2)$
    	\State $\rm img[x,y] \gets imgClean[x,y]$
    	\State $\rm prob \gets prob+0.1$
    \EndWhile
\EndIf
\end{algorithmic} 
\end{algorithm} 

\subsection{Student-Scaling}
In our investigation, we have found that scales of objects have a strong influence on object detection at night. This is due to the features of smaller objects being easily overwhelmed by glare or noise.
To allow the student to overcome this, we apply scaling augmentation to the student's inputs which includes both the image and the pseudo-labels generated by the teacher. 
As training proceeds, we follow a schedule to increase the scale of the student augmentation until it equals to that of the original image. By iteratively increasing the scale we allow the student to focus on smaller features earlier in the training process.
This process encourages the teacher to make more accurate predictions on smaller scale objects in the later stages of training. 
In turn, accurate small scale pseudo-labels allow for the increase in the scale of the student's inputs with minimal errors due to scale.

To ensure the knowledge of the previous scales is not forgotten, a gaussian function for the scaling factor is applied. The norm of the Gaussian function is obtained from the schedule values.
To prevent additional noise due to pseudo-labels being too small, labels that has an area below a threshold are removed.

\subsection{NightAug}
Night images suffer from a range of complications that are not present in daytime scenes. 
This causes a problem in the student-teacher framework, where the student would be biased towards the source domain. 
Previous methods have attempted to address this, but have either required compute-intensive translations \cite{deng2021unbiased,He_2022_CVPR} or adding additional domain classifiers to the framework \cite{li2022cross} which complicates training.
We propose NightAug, a nighttime specific augmentation pipeline that is compute-light and does not require training. 
NightAug consists of a series of augmentations with the aim of steering the characteristics of daytime images to resemble that of a nighttime image.

The defining features of nighttime images are that they are darker and have lower contrast than daytime images.
In addition the signal-to-night ratio (SNR) could be higher due to the properties of digital cameras such as luminance and colour noise. 
Glare and glow from street lamps and headlights are also present in nighttime images.
Additionally, images may be out-of-focus due to the cameras inability to detect reference points to focus on in dark environments. 

Keeping in mind the properties of nighttime images, our NightAug includes random; brightness, contrast, gamma, gaussian noise, gaussian blur augmentations and random glare insertion. 
The augmentations are randomly applied to the images and are also random in intensity. This randomness results in a wider variance of images that are exposed to the student leading to more robust training \cite{noisystudent}. 
To further increase the variance of the images, at each augmentation step, random segments of the image will ignore the application of that augmentation. 
This allows for the representation where different areas of nighttime images may be unevenly lighted. This uneven lighting affects the above characteristics of the local region.

\begin{figure}[t]
\centering
\includegraphics[width=0.48\textwidth]{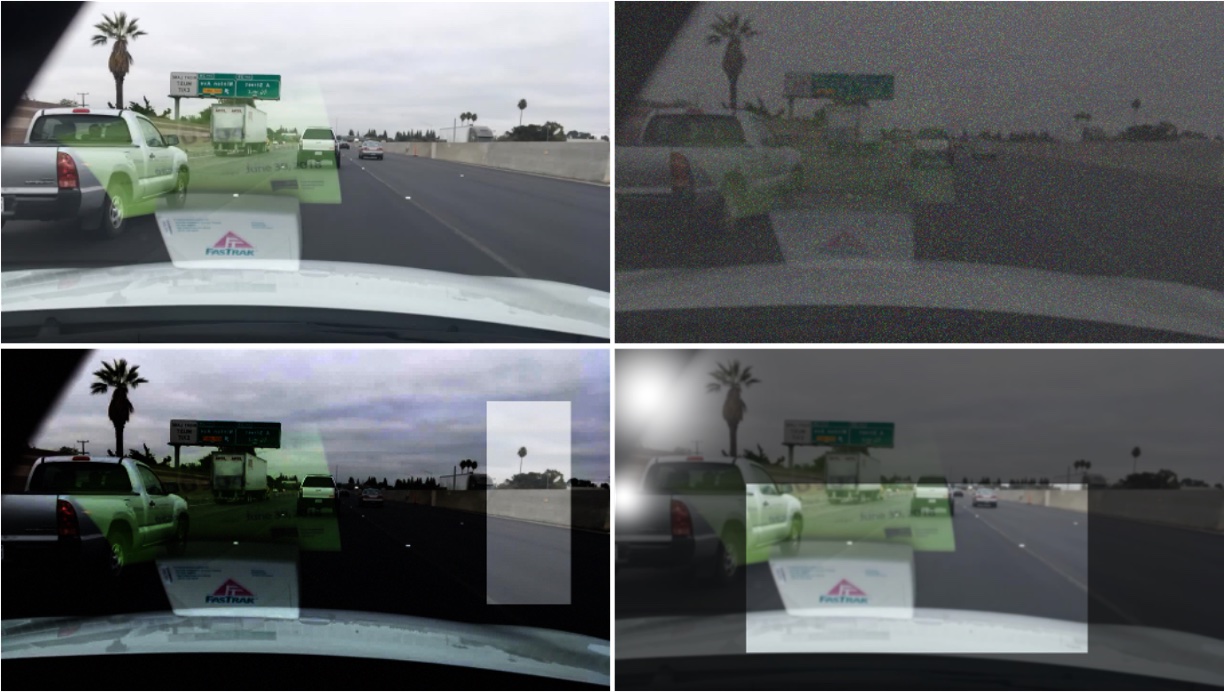}
\caption{NightAug: Original image (top-left) and images with random augmentations from: gaussian blur, gamma correction, brightness, contrast, glare, gaussian noise and random cut-outs.}
\label{fig:nightaug}
\end{figure}

A single augmentation flow of NightAug is demonstrated in Algorithm \ref{alg:cap}. Samples of an image processed with NightAug are shown in Figure \ref{fig:nightaug}. 
Each augmentation has a set probability of being applied, with the strength of the augmentation being random. 
Random regions of the augmented image may then be replaced with that of the original image. 
The probability of this region replacement reduces with each iteration. 

\vspace{-0.3cm}
\paragraph{Overall Loss}
Our total loss can be represented as:
\begin{equation}
\begin{split}
L_{\rm total} = L_{\rm sup} + \lambda L_{\rm unsup},
\end{split}
\end{equation}
where $\lambda$ represents a weight factor for the unsupervised loss, and is set experimentally. $ L_{\rm sup}, L_{\rm unsup}$ refer to Eq.~(\ref{eq:sup}) and Eq.~(\ref{eq:unsup}), respectively.

\section{Experiments}
\begin{table*}[t]
\centering
\renewcommand{\arraystretch}{1.1}
\begin{tabularx}{\textwidth}{ |l|X X X X X X X X X X|  }
 \hline
 Method & AP & Pedes-trian & Rider & Car & Truck & Bus & Motor \newline cycle &Bicycle &Traffic Light & Traffic Sign\\
 \hline
 \hline
Lower-Bound & 41.1 & 50.0 & 28.9 & 66.6 & 47.8 & 47.5 & 32.8 & 39.5 & 41.0 & 56.5 \\ 
\hline
Upper-Bound & 46.2 & 52.1 & 35.0 & 73.6 & 53.5 & 54.8 & 36.0 & 41.8 & 52.2 & 63.3\\
 \hline \hline
DA F-RCNN~\cite{chen2018domain} &  41.3 & 50.4 & 30.3 &  66.3 & 46.8 & 48.3 & 32.6 & 41.4 & 41.0 &  56.2\\
TDD~\cite{He_2022_CVPR} & 34.6 & 43.1 & 20.7 & 68.4  & 33.3  & 35.6  & 16.5 & 25.9  &  43.1 & 59.5  \\
UMT~\cite{deng2021unbiased} & 36.2 &  46.5 &  26.1 &  46.8 &  44.0 &  46.3 &  28.2 &  40.2 &  31.6 &  52.7 \\
AT~\cite{li2022cross} & 38.5 &  42.3 &  30.4 &  60.8 &  48.9 &  52.1 &  34.5 &  42.7 &  29.1 &  43.9 \\
 \hline
\textbf{2PCNet (Ours)} & {\bf 46.4} &  {\bf 54.4} &  {\bf 30.8} &  {\bf 73.1} &  {\bf 53.8} &  {\bf 55.2} &  {\bf 37.5} &  {\bf 44.5} &  {\bf 49.4} &  {\bf 65.2} \\
\hline 
 \hline
\end{tabularx}
\caption{Results of day-to-night domain adaptation on the BDD100K dataset, the Average Precision (AP) of all classes are reported. Faster RCNN detector with ResNet-50 feature extractor is used for all experiments to ensure a fair comparison. Faster RCNN is used as the lower-bound and upper-bound and is trained on labelled daytime and nighttime data respectively. The lower-bound provides a baseline without any domain adaptation while the upper-bound is fully supervised, the case where labelled target night data is available.}

\label{table:bdd100k}
\end{table*}

\begin{table}[t]
\centering
\renewcommand{\arraystretch}{1.1}
\begin{tabularx}{0.48\textwidth}{ |l|X X X X|  }
 \hline
 Method & $AP_{coco}$ & Car & Bus & Truck\\
 \hline
 \hline
Lower-Bound & 22.1 & 37.5 & 29.8 & 30.7\\ 
\hline
Upper-Bound & 23.9 & 42.0 & 33.8 & 35.0\\
 \hline \hline
FDA~\cite{fda} & 22.6 & 38.5 & 37.2 & 23.2 \\
ForkGAN~\cite{forkgan} & 22.9 & {\bf 41.2} & 33.3 & 32.1\\
 \hline
\textbf{2PCNet (Ours)} & {\bf 23.5} & 40.7 & {\bf 38.2} & {\bf 35.0}\\
\hline 
 \hline
\end{tabularx}
\caption{
	Comparison of our framework, 2PCNet, with image-to-image (I2I) translation methods. Conducted on the BDD100K dataset. ForkGan and FDA are used for comparison. Reported $AP_{coco}$ is the averaged AP over IoUs 0.5 to 0.95.
}
\label{table:bdd100k_i2i}
\vspace{-3mm}
\end{table}
\subsection{Baselines}

To evaluate our method, we compare our approach with SOTA methods in domain adaptation for object detection. 
These include DA-Faster RCNN \cite{chen2018domain}, TDD \cite{He_2022_CVPR}, UMT \cite{deng2021unbiased}, AT \cite{li2022cross} as well as a non-DA baseline Faster-RCNN \cite{frcnn}. Faster-RCNN is used as both our lower and upper-bound, where it is trained on labelled source and target data respectively. 
We additionally compare our approach with image-to-image translation methods, ForkGAN \cite{forkgan} and FDA \cite{fda}. Translation methods are trained on Faster RCNN with both the daytime and translated images.

\subsection{Datasets}
The majority of existing nighttime datasets either focuses on semantic segmentation which do not provide labels for object detection \cite{d2ndata,SDV19,SDV21}, or contains very few classes \cite{nightowls,morawski2021nod}. BDD100K \cite{bdd100k} was selected as it provides object detection labels which includes a wide range of classes (10).  It also has a large number of images compared to other DA datasets covering daytime, nighttime and other adverse conditions. 

The SHIFT \cite{shift} dataset is a recent simulated driving dataset that contains scenes in various environments. A continuous shift of these environments is available. SHIFT contains 6 class labels that share similarities to the BDD100K classes. For our evaluation, we use images with the 'day' and 'night' label as our source and target data respectively. We further ensure that the weather tag is 'clear' to isolate other weather conditions from the evaluation.

\begin{figure*}[t]
\centering
\includegraphics[width=\textwidth]{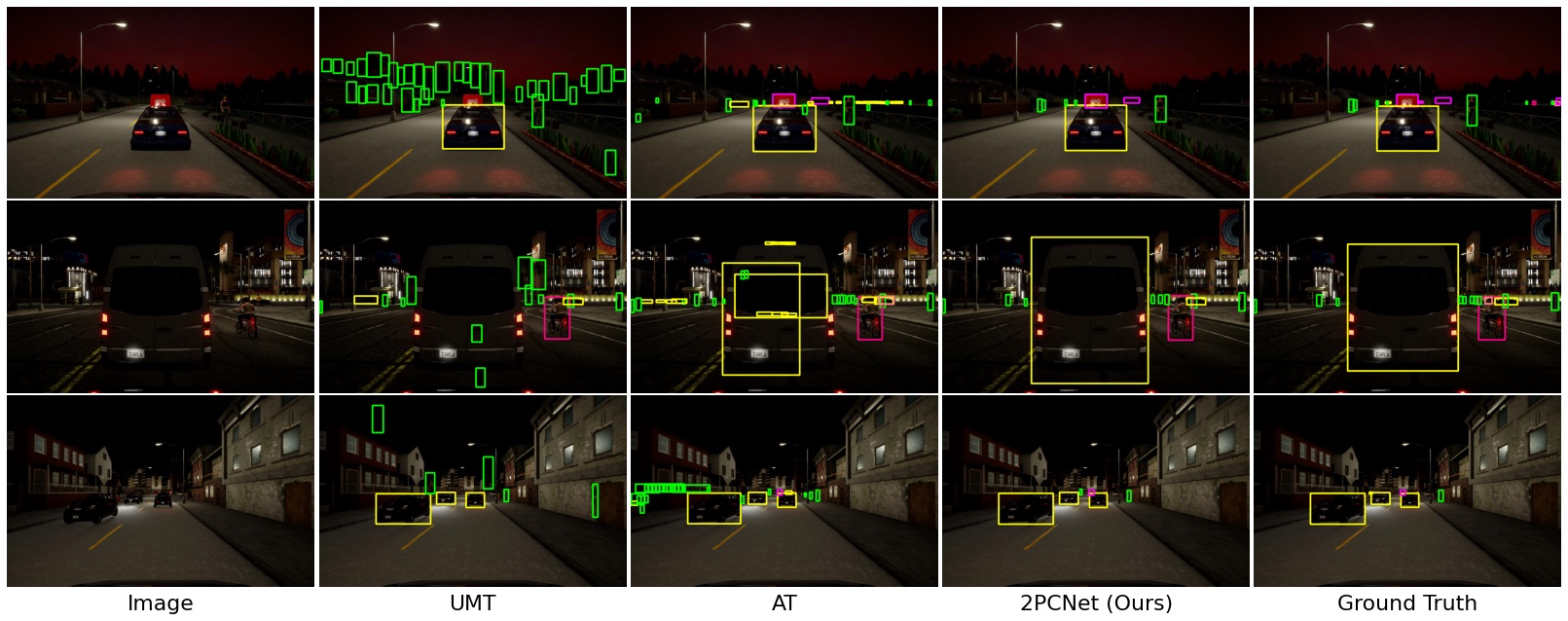}
\caption{Qualitative results of Faster RCNN, Adaptive Teacher (AT) and our method on the SHIFT dataset with the ground-truth on the far right. We can observe that Faster RCNN is not able to detect objects due to absence of domain adaptation, while AT has a large number of small false positive bounding boxes compared to our method which closely resembles that of the ground-truth.}
\label{fig:shiftqual}
\end{figure*}

\subsection{Implementation}

Following previous SOTA methods, we employ Faster-RCNN \cite{frcnn} as our base detection model and ResNet-50 \cite{resnet} pretrained on ImageNet \cite{deng2009imagenet} as our feature extractor. All images are scaled by resizing its shorter side to 600 pixels. For student-scaling we set a schedule for (0.57, 0.64, 0.71, 0.78, 0.85, 0.92) of the maximum iterations at scales (0.5, 0.6, 0.7, 0.8, 0.9, 1.0). Loss hyperparameters are set at $\lambda\ = 0.3$ and the rate smooth coefficient parameter of the EMA is $0.9996$. A confidence threshold of $0.8$ for phase one of Two-Phase Consistency. For the initial pretraining of the student model, we train the student for 50k and 20k iterations on the source images, for BDD100K and SHIFT respectively. Supervised inputs are daytime images with and without NightAug. We then copy the weights to the teacher and continue training with the addition of unsupervised loss for an additional 50k iterations. The learning rate is kept at 0.04 throughout training. Our network is trained on 3 RTX3090 GPUs with a batch-size of 6 source and 6 target images.

\subsection{Comparison to SOTA}
\paragraph{Comparison on BDD100K} We compare our method against the SOTA on real driving scenes and evaluating their domain adaptation performance on nighttime images, the results of this experiment can be seen on Table \ref{table:bdd100k}.
The results show that our method achieves the highest performance with an AP of 46.4. 20.5\% higher than that of the SOTA student-teacher methods and above that of the upper-bound. 
We have observed in experiments that student-teacher methods underperforms with an AP below that of the lower-bound due to the error-propagation from noisy pseudo-labels. The result of the error is small false positive detections as seen in Figure \ref{fig:bdd100kqual}. Our method does not suffer from the same allowing for higher performance.
\begin{table}[t]
\small
\centering
\renewcommand{\arraystretch}{1.1}
\begin{tabularx}{0.48\textwidth}{ |l|X X X X X X X|  }
 \hline
 Method & AP & Per. & Car & Truck & Bus & Mcy. & Bcy. \\
 \hline
 \hline
Lower-Bound & 41.6 &  40.4 &  44.5 &  49.9 &  53.7 & 14.3  & 46.7  \\
Upper-Bound & 47.0 &  49.7 &  51.5 &  56.0 &  53.6 & 19.2  & 52.4  \\
 \hline  \hline
DA FR~\cite{chen2018domain} & 43.7 &  43.0 &  48.8 &  47.8 & 52.1 & 19.9  & {\bf 55.8}  \\
UMT~\cite{deng2021unbiased} & 31.1 &  7.7 &  47.5 &  18.4 &  46.8 & 16.6  & 49.2  \\
AT~\cite{li2022cross} & 38.9 & 25.8 &  33.0 &  54.7 &  49.5 & 20.7  & 52.3  \\
 \hline
\textbf{2PCNet (Ours)} & {\bf 49.1} &  {\bf 51.4} &  {\bf 54.6} & {\bf 54.8} & {\bf 56.6} & {\bf 23.9}  & 54.2 \\
\hline  
 \hline 
\end{tabularx}
\caption{Results of Day-to-Night domain adaptation on the SHIFT dataset. The Average Precision (AP) of all classes. Faster RCNN is used as the lower-bound and upper-bound and is trained on labelled daytime and nighttime data respectively.}
\label{table:SHIFT}
\vspace{-3mm}
\end{table}
We can also observe that our method performs well across all classes. Even when compared with the upper-bound, 2PCNet achieves higher AP on the majority of classes. This indicates that our method is able to generalise well across large and small classes. 

\begin{figure*}[t]
	\centering
	\includegraphics[width=\textwidth]{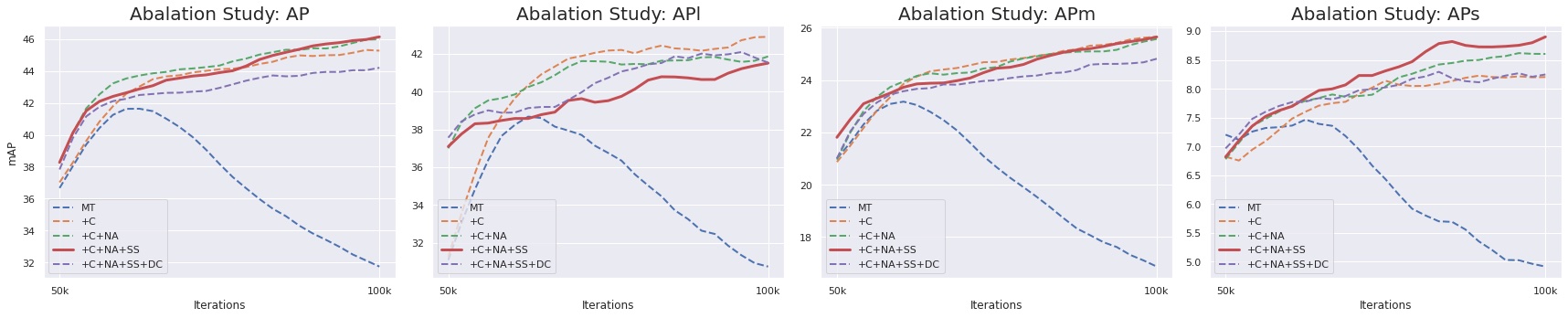}
	\caption{Training curve on BDD100K dataset ablation study. We show the overall AP training curve as well as the AP of large, medium and small objects. MT represents the base Mean Teacher framework. It can be seen that at all scales, the absence of Two-Phase Consistency (C) results in a sharp drop during training. We can also see that with the inclusion of NightAug (NA) and student-scaling (SS) the gradient of the curve increases. We note that the inclusion of a domain classifier (DC) reduces the performance at all scales.}
	\label{fig:aba_graphs}
\end{figure*}

The comparison with image-to-image translation methods is shown in Table \ref{table:bdd100k_i2i}. Translation methods do not suffer from the error propagation problem as it is trained on Faster RCNN without a teacher. Even so, we can see that our method outperforms SOTA adverse vision translation methods. 
\vspace{-0.3cm}
\paragraph{Comparison on SHIFT} To further compare our method with SOTA we evaluate on the SHIFT simulation dataset. Due to the nature of the simulated data, many nighttime image characteristics that we have previously mention is not exhibited in this data such as blurriness, noise and glare. 

The results of this experiments are shown in Table \ref{table:SHIFT}. 
We can observe that previous SOTA methods that use the student-teacher framework perform worse than the lower-bound. The sub-par performance is again due to the error-propagation problem. AT performs better than UMT due to ATs inclusion of adversarial learning.
However, adversarial learning is not enough to mitigate this problem. 
We can see that the performance of DA FRCNN outperforms both the SOTA student-teacher methods as it would not be affected by error-propagation. It is however, still largely below the upper-bound performance. 
2PCNet outperforms these previous methods as well as the upperbound. We achieve an improvement of +10.2 AP over previous SOTA student-teacher methods and +2.1 AP over that of the upper-bound. 

\subsection{Ablation Studies}
To demonstrate the effectiveness of each of our components, we train several models for 100K iterations and evaluate them on the BDD100K dataset. We present our findings in Table \ref{table:ablation}. 

\vspace{-0.3cm}
\paragraph{Two-Phase Consistency} We can observe in Table \ref{table:ablation} that the addition of Two-Phase Consistency (C) demonstrated a wide performance gap when compared to the Mean-Teacher baseline, +13.5 AP (43\%). This improvement in AP exists across large, medium and small objects. While the performance of MT is initially strong, it rapidly begins to decline; which can be observed in Figure \ref{fig:aba_graphs}. This drop in performance is due to the error propagation of noisy pseudo-labels. The experimental results show that Two-Phase Consistency is able to provide a solution. This ensures that highly confident pseudo-labels are bounded by less confident pseudo-label enabling a balance of knowledge into the student. 

\vspace{-0.3cm}
\paragraph{NightAug} We benched marked the effectiveness of NightAug in our framework as shown in  Table \ref{table:ablation}. The inclusion of NightAug increases the detection performance of small objects with an increase of 5\%. Additionally, the gradient of the training performance remains steep as seen in Figure \ref{fig:aba_graphs}. The positive gradient is displayed most strongly for APm and APs where objects are more prone to nighttime specific complications.

\vspace{-0.3cm}
\paragraph{Student-Scaling} Our final component, student-scaling, is included into the framework and the results can be seen in Table \ref{table:ablation}. We can observe that student-scaling is able to boost the performance of small object detection by 6\%. This boost in performance is due to the student network focusing on smaller object earlier in the training process. We note that the performance of large objects have dropped by 1-2\%; however when referring to the training curves in Figure \ref{fig:aba_graphs}, APl remains steep. As the initial focus is on smaller objects, less time is allocated to larger objects during training. This can be mitigated by lengthening training resulting in more iterations for larger objects.

\vspace{-0.3cm}
\paragraph{Domain Classifier} To conclude our study, we included a domain classifier into our network. Adversarial learning is a widely used DA technique; however when added into 2PCNet, a performance drop across all scales can be seen. This drop is shown in Table \ref{table:ablation}. The suppression of nighttime features is suspected to be the cause. Suppression is present as the adversarial loss guides the feature extractor to maintain domain invariancy. By suppressing nighttime features, the teacher has less information to distil to the student. 
This is demonstrated in Figure \ref{fig:aba_graphs} where the domain classifier (dotted purple) initially performs well. But as training continues, our method (solid red) is able to surpass its performance.

\begin{table}[t]
\centering
\renewcommand{\arraystretch}{1.1}
\begin{tabularx}{0.45\textwidth}{ |c c c c |X |X| X |X|  }
 \hline
 \multicolumn{4}{|c|}{Methods} &  \multicolumn{4}{c|}{}\\
 
 C & NA & SS & DC & {\small AP} & {\small APl} & {\small APm} & {\small APs} \\
\hline 
\hline 
\checkmark &  \checkmark & \checkmark &&46.4 & 41.7 & 25.8 & 9.1\\
\hline 
\checkmark &  \checkmark & \checkmark &\checkmark &44.5 & 41.6 & 25.0 & 8.3\\
\checkmark &  \checkmark & && 45.8 & 42.2 & 25.7 & 8.6\\ 
\checkmark &  && & 45.2 & 42.9 & 25.7 & 8.2\\
&  & && 31.7 & 30.4 & 16.5 & 4.8\\
\hline 
\end{tabularx}
\caption{Ablation studies on the BDD100K dataset. The last row represents the base Mean-Teacher network. Methods are referred to as, C: Two-Phase Consistency, NA: NightAug, SS: Student-Scaling, DC: Domain Classifier. APl, APm, and APs represent the AP of large, medium and small objects respectively.}
\label{table:ablation}
\vspace{-3mm}
\end{table}

\section{Conclusion}
Our proposed framework, 2PCNet, presents a novel solution to the challenges of day-to-night domain adaptive object detection. With our Two-Phase Consistency approach, we are able to effectively leverage high and low confidence knowledge for the student, while mitigating error propagation commonly present in previous student-teacher methods. We further address issues arising from small scale and dark objects through the use of student-scaling and NightAug, respectively. Experimental results on the e BDD100K \cite{bdd100k} and SHIFT \cite{shift} datasets demonstrate that 2PCNet outperforms existing state-of-the-art methods. Overall, our proposed framework provides an effective and efficient solution for day-to-night domain adaptive object detection.
\paragraph{Acknowledgements} This work is partially supported by MOE2019-T2-1-130.

{\small
\bibliographystyle{ieee_fullname}
\bibliography{v3_mikhail}
}

\end{document}